\pdfoutput=1
\documentclass[11pt]{article}

\usepackage{tracefnt}

\usepackage[final]{acl}

\usepackage{times}
\usepackage{latexsym}

\usepackage[T1]{fontenc}
\usepackage[utf8]{inputenc}

\usepackage{microtype}

\usepackage{inconsolata}

\usepackage{graphicx}

\usepackage{amsmath}
\usepackage{amsfonts}
\usepackage{booktabs}
\usepackage{multirow}

\usepackage{listings}
\usepackage{xcolor}

\definecolor{codegray}{rgb}{0.5,0.5,0.5}
\definecolor{codepurple}{rgb}{0.58,0,0.82}
\definecolor{backcolor}{rgb}{0.95,0.95,0.95}

\lstdefinestyle{mypython}{
  backgroundcolor=\color{backcolor},   % 背景色
  commentstyle=\color{codegray}\ttfamily,
  keywordstyle=\color{blue}\bfseries,
  numberstyle=\tiny\color{gray},
  stringstyle=\color{codepurple},
  basicstyle=\ttfamily\footnotesize,   % 基本字体
  breakatwhitespace=false,             
  breaklines=true,                     
  captionpos=b,                        
  keepspaces=true,                     
  numbers=left,                        
  numbersep=5pt,                       
  showspaces=false,                    
  showstringspaces=false,
  showtabs=false,                      
  tabsize=4,
  frame=single,                        % 添加边框
  rulecolor=\color{gray},             % 边框颜色
  language=Python
}

\title{KV-Latent: Dimensional-level KV Cache Reduction with Frequency-aware \\Rotary Positional Embedding}

\author{
 \textbf{Luohe Shi\textsuperscript{1}},
 \textbf{Zuchao Li\textsuperscript{2}\footnotemark[1]},
 \textbf{Lefei Zhang\textsuperscript{1}},
 \textbf{Guoming Liu\textsuperscript{3}},
 \textbf{Baoyuan Qi\textsuperscript{3}},
 \textbf{Hai Zhao\textsuperscript{4}}
\\ \\
 \textsuperscript{1}School of Computer Science, Wuhan University, Wuhan, China \\
 \textsuperscript{2}School of Artificial Intelligence, Wuhan University, Wuhan, China;
 \textsuperscript{3}Xiaomi, Beijing, China \\
 \textsuperscript{4}School of Computer Science, Shanghai Jiao Tong University, Shanghai, China \\
 \texttt{\{shiuohe, zcli-charlie, zhanglefei\}@whu.edu.cn} \\
 \texttt{\{qibaoyuan, liuguomin\}@xiaomi.com} \texttt{zhaohai@cs.sjtu.edu.cn} \\
\\
}

\begin{document}

\maketitle

\begin{abstract}
Large language models (LLMs) based on Transformer Decoders have become the preferred choice for conversational generative AI. Despite the overall superiority of the Decoder architecture, the gradually increasing Key-Value (KV) cache during inference has emerged as a primary efficiency bottleneck, both in aspects of memory consumption and data transfer bandwidth limitations. To address these challenges, we propose a paradigm called KV-Latent. By down-sampling the Key-Value vector dimensions into a latent space, we can significantly reduce the KV Cache footprint and improve inference speed, only with a small amount of extra training, less than 1\% of pre-training takes. Besides, we enhanced the stability of Rotary Positional Embedding applied on lower-dimensional vectors by modifying its frequency sampling mechanism, avoiding noise introduced by higher frequencies while retaining position attenuation. Our experiments, including both models with Grouped Query Attention and those without, have yielded satisfactory results. Finally, we conducted comparative experiments to study the impact of separately reducing Key and Value components on model's performance. Our approach allows for the construction of more efficient language model systems, and opens the new possibility on KV Cache saving and efficient LLMs. Our code is available at \href{https://github.com/ShiLuohe/KV-Latent}{https://github.com/ShiLuohe/KV-Latent}.
\end{abstract}


\renewcommand{\thefootnote}{\fnsymbol{footnote}}
\footnotetext[1]{Corresponding Author. This work was supported by the National Natural Science Foundation of China (No. 62306216), the Technology Innovation Program of Hubei Province (No. 2024BAB043) and Xiaomi Open-Competition Research Program.}

\section{Introduction}

The release of ChatGPT~\citep{NEURIPS2020_1457c0d6} launched an generative AI trend, and as the core architecture behind these state-of-the-art models, the Transformer decoder~\citep{NIPS2017_3f5ee243} has gain many attention. There capabilities are spectacular~\citep{DBLP:journals/corr/abs-2312-11805, tang-etal-2025-nota, anthropic2024claude3, anthropic2025claude4, yang-etal-2025-label, DBLP:conf/nips/ShiYLZ024}. Undeniably, as large language models (LLMs) become more integrated into people’s lives, the costs associated with training and inference are increasingly impossible to ignore~\citep{DBLP:conf/prcv/YanLZ24}. While training costs remain relatively fixed and centralized, inference costs grow linearly with user adoption and are often distributed across different spaces and timeframes, making the optimization of model inference costs increasingly urgent. The Transformer decoder architecture, employed by LLMs, operates as a causal model, avoiding the need to recompute most intermediate states during a autoregressive generation. However, it still requires retaining certain intermediate states. Specifically, as a self-attention-based architecture, it necessitates preserving the key and value (KV) states corresponding to each token, and commonly referred to as the KV Cache. 
% The KV Cache size increases with the number of tokens. Furthermore, because each dialogue request maintains its own dedicated KV Cache, accelerate the system with batch processing is impossible, leading to RAM bandwidth bottlenecks and wasted computational resources on chips.
The time complexity of the self-attention mechanism is uniformly $O(n^2)$, meaning that for each additional token in a sequence, the computational workload increases at least by $O\left((n^2)'\right)=O(n)$. Consequently, in typical situations, we need to interact with $O(n)$ cached states. In other words, the required storage for the KV Cache grows linearly with the generation of tokens. This poses a significant challenge.

% Despite this, a complete transition to linear models~\citep{pmlr-v119-katharopoulos20a} such as RWKV~\citep{DBLP:conf/emnlp/PengAAAABCCCDDG23} or RetNet~\citep{DBLP:journals/corr/abs-2307-08621} remains elusive. These models only require interaction with a fixed number of states for each token generated. However, we cannot fit the linearly increasing contextual information into this limited size, regardless of how sophisticated our methods are. Therefore, if we aspire to achieve the dominant performance of current Transformers in handling long texts, including challenges like Needle in a haystack~\citep{DBLP:journals/corr/abs-2308-14508, NEURIPS2023_ab05dc8b}, we must accept the existing complexity.

The KV Cache faces two primary challenges: growing volume and non-friendly access pattern. The large volume necessitates increasingly expensive hardware for efficient KV Cache storage and retrieval, furthermore, because each inference request maintains its own dedicated KV Cache, accelerate the system with batch processing is impossible, leading to RAM bandwidth bottlenecks and wasted computational resources on chips~\citep{10.1145/1498765.1498785}. Meanwhile, the non-friendly memory access arises due to the cache size frequently fluctuating. The latter challenge can be significantly mitigated through more scientifically organized cache structures, such as \textit{paged attention}~\citep{10.1145/3600006.3613165} or heterogeneous inference systems like \textit{fastdecode}~\citep{he2024fastdecodehighthroughputgpuefficientllm}. However, the former challenge remains more intricate.

To address the issue of KV Cache size, several methods have been proposed. At attention-head-level, Multi-Query Attention (MQA, \citealp{DBLP:journals/corr/abs-1911-02150}), Grouped Query Attention (GQA, \citealp{DBLP:conf/emnlp/AinslieLJZLS23}) are effective and widely proved methods. At layer-level, cross-layer reuse methods has been proposed, such as \textit{You Only Cache Once}~\citep{DBLP:journals/corr/abs-2405-05254} and \textit{Cross Layer Attention}~\citep{DBLP:journals/corr/abs-2405-12981}. At token-level, researchers have focused on eviction and merging, methods include \textit{Heavy Hitter Oracle}~\cite{NEURIPS2023_6ceefa7b}, \textit{PyramidInfer}~\cite{DBLP:journals/corr/abs-2405-12532}, \textit{SirLLM}~\cite{DBLP:journals/corr/abs-2405-12528}, and $L_2$ Norm method proposed by \citeauthor{DBLP:journals/corr/abs-2406-11430}.

% Multi-Query Attention (MQA), proposed by \citeauthor{DBLP:journals/corr/abs-1911-02150}, as a pioneering work in this field, introduced the concept of intra-layer KV Cache reuse. MQA combines all \texttt{Key} and \texttt{Value} heads into two single heads and queries the single \texttt{Key} head multiple times using various \texttt{Query} heads, resulting in different weight combinations for merging \texttt{Value} heads. Building upon this, Grouped Query Attention (GQA), proposed by \citeauthor{DBLP:conf/emnlp/AinslieLJZLS23}, further expands MQA’s approach. Instead of merging all \texttt{Key} and \texttt{Value} heads into one, GQA pre-groups all attention heads. Within each group, multiple \texttt{Query} heads share a single \texttt{Key} head, generating weighted combinations for the correspondingly single \texttt{Value} heads. 
% Intra-layer sharing not only reduces the KV Cache size by concentrating repeated content access in time but also enhances the computational intensity of the KV Cache. This improvement aligns with \citeauthor{10.1145/1498765.1498785}’s theoretical framework, enhancing overall system efficiency.  

Despite the substantial progress made by previous research, directly reducing the size of \texttt{Key} and \texttt{Value} heads remains a less explored area. In the context of MHA, each attention head is a combination of two low-rank transformations, the first is the pair of $K$ and $Q^\top$, the second is the pair of $V$ and $O$. We define dimension of each attention head is $d_h$, the number of heads is $n_h$, and the model’s hidden dimension is $d$. We observe that $K$ and $V$ represent two linear transformations that downsample $d$-dimensional hidden state $h$ to two $d_h$-dimensional vector $k$ and $v$. Correspondingly, $Q^\top$ and $O$ perform up-sampling from $d_h$ to $d$ dimensions. The KV Cache stores the latent vectors resulting from these two low-rank transformations.

Typically, we assume that $d_h \times n_h = h$, but when considering an individual head, $d_h$ and $h$ do not necessarily need to adhere to this predefined relationship. The work of MQA and GQA, and other recent works~\citep{DBLP:journals/corr/abs-2406-07056, deepseekai2024deepseekv2strongeconomicalefficient, saxena2024eigenattentionattentionlowrank}, has already demonstrated that the retained KV Cache does not require complete $d$-dimensional vectors; low-rank representations suffice for transmitting information between tokens. However, we aim to go further by decoupling the constraint $d_h * n_h = h$. Our approach involves directly reducing the head size from existing models, then restore model's performance by a minimal amount of additional training with a 2-stage strategy, achieving the goal of KV Cache reduction. Since we essentially map the \texttt{Key} and \texttt{Value} into a latent space then directly decode from it by \texttt{Query}-transpose and \texttt{Output}, we name our proposed method KV-Latent.

Furthermore, we observe that even within individual attention heads, the low-rank transformations of $KQ^\top$ and $VO$ do not necessarily require the same dimension. Specifically, we can separate $d_h$ into $d_{qk}$ and $d_{vo}$, and these dimensions need not be equal. Building upon this insight, we explore various reduction strategies with different value of $d_{qk}$ and $d_{vo}$, to investigate their impact on training time, inference efficiency, and, the most important aspect, model’s capabilities.

Lastly, in our experiments, we discovered that the stability of Rotary Position Embedding (RoPE, \citeauthor{SU2024127063}) diminishes at lower dimensions, affecting long-range ability. By analyzing RoPE’s sampling mechanism, we identified that noise from high-frequency features dominate when the dimensionality is low. Consequently, we refined our approach by modifying RoPE’s sampling method in a frequency-aware way to maintain stability even at lower dimensions.

Out contribution includes:
\begin{itemize}
    \item We've proved that by a small amount of additional training with 2-stage strategy, we can fit the KV Cache into a latent space, thus directly reduces the space occupation and bandwidth requirement of KV Cache.
    \item By using different combinations of $d_{qk}$ and $d_{vo}$, we observed that model's performance is more sensitive to $d_{vo}$ comparing to $d_{qk}$, which reveals how LLMs are affect by different parts of self-attention, providing insights to optimize LLMs' model structure.
    \item By modifying RoPE's sampling mechanism in a frequency-aware way, excluding high frequency portions and amplifying low frequency portions, we successfully make RoPE more stable when applied on lower-dimensional \texttt{Query} and \texttt{Key}.
\end{itemize}

\section{Backgrounds and Related Works}

\subsection{Transformer Decoder}
% In the context of decoder-only models, the Transformer decoder block~\cite{NIPS2017_3f5ee243} consists of two main components: masked self-attention and feed-forward network. Each of these components has its own separate normalization layer and residual connection. 
Our primary focus lies on the masked self-attention of Transformer~\cite{NIPS2017_3f5ee243} decoder.
We define $h$ as the hidden vector of the input at $l$-th layer, token-wise, and $H$ for the whole sequence. Our goal is to compute $h'$, which represents the output of the attention module. The process is described by Formula~\ref{for:sat}, where $W^{(i)}_{\{Q,K,V,O\}}$ corresponds to the parameter matrices for the $Q$, $K$, $V$, and $O$ transformations of the $i$-th head. And the $\mathcal{K}^{(i)}, \mathcal{V}^{(i)}$ represents the KV Cache of the $i$-th head. We apply right multiplication in this context.
\begin{equation}
\begin{split}
\label{for:sat}
    & h' = \sum_{i=1}^{n_h}\left[
        \mathrm{softmax}\left(
            \frac{
                q^{(i)} \mathcal{K}^{(i)^{\top}}
            }{
                \sqrt{d_{qk}}
            }
        \right) 
        \mathcal{V}^{(i)} {W^{(i)}_O} 
    \right]\\
    & q^{(i)} = h W^{(i)}_Q,\ 
    \mathcal{K}^{(i)} = H W^{(i)}_K,\ 
    \mathcal{V}^{(i)} = H W^{(i)}_V \  
\end{split}
\end{equation}

% From Formula~\ref{for:sat}, we observe that the relationship that $d_h \times n_g = d$ is not necessary. Similarly, we notice that the dimensions of the two sets of low-rank transformations, $KQ^\dagger$ and $VO$, do not have to be the same. This implies that we can achieve KV Cache compression without significantly altering the model structure. These observations directly motivate our approach: directly reducing $d_h$ and splitting it into non-identical $d_{qk}$ and $d_{vo}$ components.

\subsection{KV Cache Reduction Methods}

\subsubsection{Head-level}
MQA\cite{DBLP:journals/corr/abs-1911-02150} combines all \texttt{Key} and \texttt{Value} heads into two single heads and queries the single \texttt{Key} head multiple times using various \texttt{Query} heads. Building upon this, GQA~\cite{DBLP:conf/emnlp/AinslieLJZLS23} pre-groups all attention heads. Within each group, multiple \texttt{Query} heads share a single \texttt{Key} head and correspondingly single \texttt{Value} heads. GQA introduces a tunable variable, the number of groups $n_g$ and the corresponding number of heads within each group, finding a new trade-off method between MQA and MHA (Multi-Head Attention). This approach provides a fine-grained balance between efficiency and performance, boosts the operational intensity, and has been widely adopted in models like LLaMA2~\citep{DBLP:journals/corr/abs-2307-09288}, LLaMA3~\citep{dubey2024llama3herdmodels}, Mistral~\citep{DBLP:journals/corr/abs-2310-06825, DBLP:journals/corr/abs-2401-04088}, and Qwen~\citep{yang2024qwen2technicalreport}. These works have proved the low-rank nature of KV Cache, which guaranteed the effectiveness of our method. 

% KV Cache, through intra-layer reuse, not only reduces it's sheer size, but also leverages the property of repeated use of the same variables in adjacent time steps. Its excellent temporal locality allows for acceleration through multi-level hardware caching, thereby enhancing computational intensity and improving system efficiency. We hope to maintain these supremacy in our methods.

\subsubsection{Layer-level}
Cross-layer reuse is another hot topic. Methods like \textit{You Only Cache Once}~\citep{DBLP:journals/corr/abs-2405-05254} and \textit{Cross Layer Attention}~\citep{DBLP:journals/corr/abs-2405-12981} have successfully reduced KV Cache size by reusing the same KV Cache states across different decoder layer. However, Due to the non-continuous nature of reused content over time, cross-layer reuse cannot optimize computational intensity effectively, and bandwidth limitations from data exchanges persist, limiting inference speed improvement. 

\subsubsection{Token-level}
In token level, eviction~\cite{NEURIPS2023_a452a7c6} and merging~\cite{DBLP:journals/corr/abs-2402-07616, anonymous2025spindlekv} are the most essential methods, for which the core idea is to evict less attend tokens or to merge states from multiple tokens into one. Popular works includes \textit{Heavy Hitter Oracle}~\cite{NEURIPS2023_6ceefa7b}, \textit{PyramidInfer}~\cite{DBLP:journals/corr/abs-2405-12532}, \textit{SirLLM}~\cite{DBLP:journals/corr/abs-2405-12528}, and $L_2$ Norm method proposed by \citeauthor{DBLP:journals/corr/abs-2406-11430}.
Possible problem for token level reduction lies in the reliance on the attention score, making them cannot be combined with prefill acceleration methods, for example \textit{flash attention}~\cite{NEURIPS2022_67d57c32}. Other methods that do not rely on attention often lacks fine granularity, risking critical information loss. Achieving practical large-scale usage remains a challenge.

% \subsubsection{Other methods} Other reduction methods includes:

% \textbf{Quantization}~\cite{DBLP:journals/corr/abs-2401-18079, zhang2024pqcacheproductquantizationbasedkvcache} of KV Cache has been proposed. But the online-nature of KV Cache make most of the mature but offline quantization methods unable to be applied on.

% \textbf{Multi-head Latent Attention}~\cite{deepseekai2024deepseekv2strongeconomicalefficient}, directly maps the KV Cache to a latent space, effectively reducing its size. 
% For MLA, its weakness, that it requires to expand latent vector into higher dimensional space eventually, still ties it to data exchange bandwidth limitations.  

\subsection{Rotary Positional Embedding}

Rotary Position Embedding (RoPE), proposed by \citeauthor{SU2024127063} in 2021, is a method that enhances position encoding for Decoder models. This type of position encoding has gained widespread adoption due to its various desirable properties. First, it adheres to the characteristic of long-range attenuation: the farther apart two identical vectors are in a sequence, the weaker their attention connection becomes. Second, RoPE is a form of relative position encoding, meaning that the attenuation remains consistent for the same relative distances. This property contributes to better generalization. Finally, RoPE achieves its encoding through sparse matrices, resulting in computational efficiency. These favorable properties make it nearly the sole choice for modern LLMs. However, our experiments revealed that RoPE exhibits instability at lower dimensions due to high periodic noise. We mitigated this issue by modifying its frequency sampling approach.

\section{Methods}

\subsection{Preliminary and Notations}

Applying RoPE to Formula~\ref{for:sat}, we achieve Formula~\ref{for:satwRoPE}. 
\begin{equation}
\begin{split}
\label{for:satwRoPE}
    & h' = \sum_{i=1}^{n_h}\left[
        \mathrm{softmax}\left(
            \frac{
                q^{(i)}\mathcal{R}^{\theta,\delta}_{x}\mathcal{K}^{(i)^{\top}}
            }{
                \sqrt{d_{qk}}
            }
        \right) 
        \mathcal{V}^{(i)} {W^{(i)}_O} 
    \right]\\
    & q^{(i)} = h W^{(i)}_Q,\ 
    \mathcal{K}^{(i)} = H W^{(i)}_K \mathcal{R},\  
    \mathcal{V}^{(i)} = H W^{(i)}_V\ 
\end{split}
\end{equation}
In which $h$ refers to the hidden state of a single token, correspondingly, $H$ as the hidden states of the whole sequence. The parameter of four linear transformation of self-attention is given by $W_{\{Q,K,V,O\}}$, and the transformation here is in the form of right matrix multiplication. 
We define $d_{qk}$ as the dimension of each \texttt{Query} and \texttt{Key} head, and $d_{vo}$ as \texttt{Value} and \texttt{Output} head here. 
With $n_h$ as the amount of heads, we can get $W_{\{Q,K\}}\in \mathbb{R}^{d\times n_hd_{qk}}$ and $W_{V}\in \mathbb{R}^{d\times n_hd_{vo}},\ W_{O}\in \mathbb{R}^{n_hd_{vo} \times d}$. In this case, we define $W_{\{Q,K,V,O\}}^{(i)}$ as the parameter that corresponds to the $i$-th head, $i\in[1,2,\dots,n_h]$, as Formula~\ref{for:headsplit}.
\begin{equation}
\label{for:headsplit}
\begin{split}
    W_Q^{(i)} & = W_{Q}[\,:\,, (i-1)d_{qk}:id_{qk}] \in \mathbb{R}^{d \times d_{qk}} \\
    W_K^{(i)} & = W_{K}[\,:\,, (i-1)d_{qk}:id_{qk}] \in \mathbb{R}^{d \times d_{qk}} \\
    W_V^{(i)} & = W_{V}[\,:\,, (i-1)d_{vo}:id_{vo}] \in \mathbb{R}^{d \times d_{vo}} \\
    W_O^{(i)} & = W_{O}[(i-1)d_{vo}:id_{vo}, \,:\,] \in \mathbb{R}^{d_{vo} \times d}
\end{split}
\end{equation}

We additional define $x$ as the position of current token, $\mathcal{R}_{\theta, \delta}(x)$ as the rotary matrix defined in RoPE for the $x$-th position, $\delta=\frac d2$. More precisely, according to RoPE, $\mathcal{R}$ is given out in Formula~\ref{for:ropedefault}.
\begin{equation}
\label{for:ropedefault}
\begin{split}
    & \mathcal{R}_{\theta, \delta}(x) =
    \begin{bmatrix}
        \mathbf{R}_{\theta,1}(x) & 0 & \dots & 0 \\
        0 & \mathbf{R}_{\theta,2}(x) & \dots & 0 \\
        \vdots & \vdots & \ddots & \vdots \\
        0 & 0 & \dots & \mathbf{R}_{\theta,\delta}(x) \\
    \end{bmatrix}
    \\
    & \mathbf{R}_{\theta,j}(x) = \begin{bmatrix}
        \cos x\theta_j & -\sin x\theta_j \\
        \sin x\theta_j &  \cos x\theta_j
    \end{bmatrix}\!,\ \theta_j = \theta^{-\left(j-1\right)/ \delta}
\end{split}
\end{equation}

\subsection{KV-Latent with Two-Stage Training}
We propose the KV-Latent paradigm, which aims to reduce KV Cache by directly modifying the shape of pre-trained model’s $W_K$ and $W_V$. Subsequently, we restore model performance through fine-tuning with a smaller amount of data. The paradigm involves a RoPE compatible attention down-sampling strategy and a two-stage continuation training.

\subsubsection{Model Preparation}

% To achieve KV-latent vectorization, we need a down-sampling process for the KV pairs. However, we aim to minimize the loss of the model’s initial capabilities during this process. The straightforward approach involves randomly selecting a desirable amount of dimensions within each attention head. Without considering RoPE, this method works well because any dimension within the same head satisfies rotational symmetry. However, when accounting for the RoPE, we need to reconsider our approach. 

Before training, we need to initialize a copy of the model after dimensionality reduction of the attention heads~\citep{DBLP:journals/corr/abs-2503-23924}. For any given attention model, random sampling alone is sufficient to retain the information in the attention matrix in an adequately balanced manner, as the channels within each attention head exhibit rotational symmetry. This means that it suffices to preserve the same channels for both $QK^\top$ and $VO$. 

\begin{figure}[b]
    \centering
    \includegraphics[width=1\linewidth]{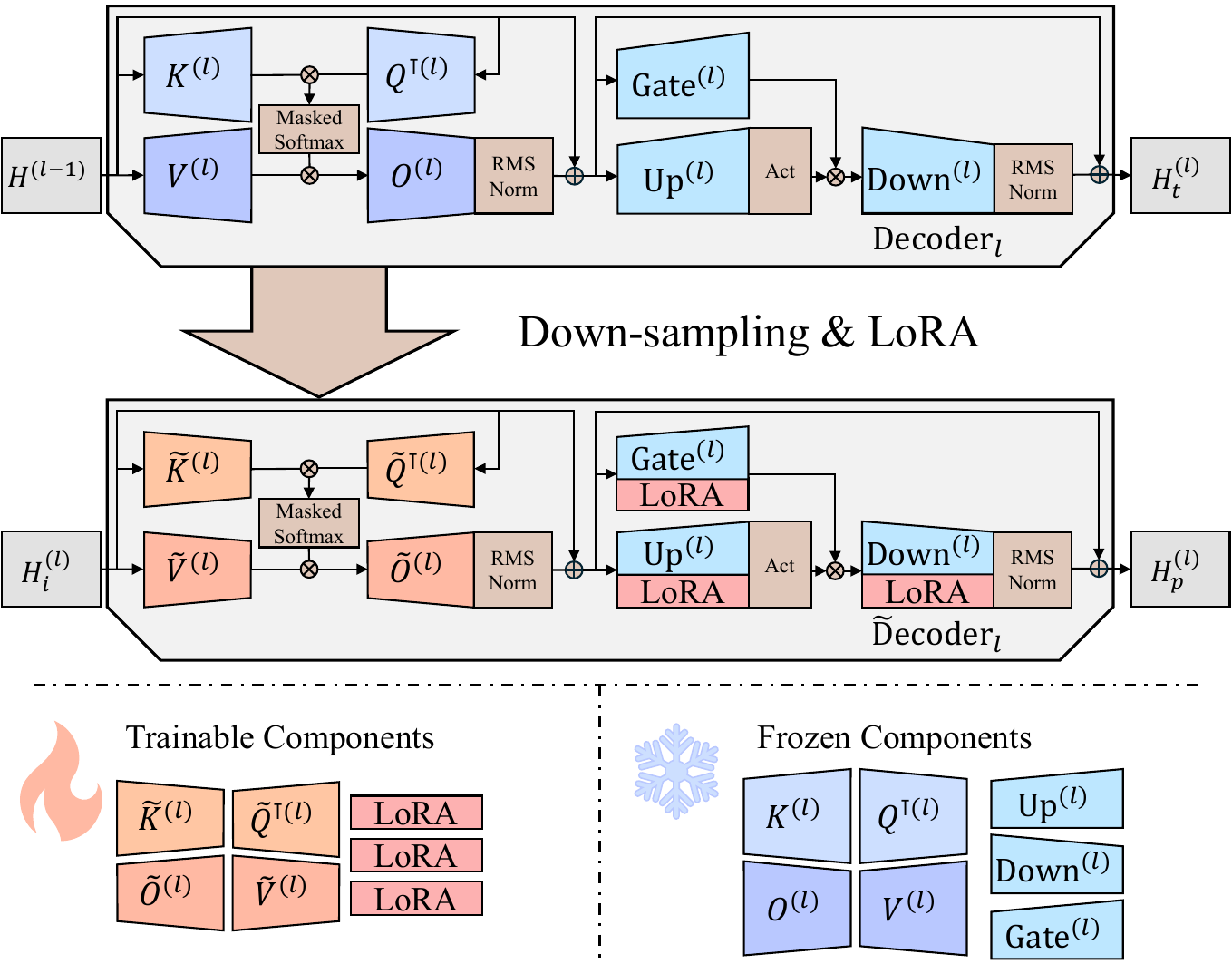}
    \caption{Model preparation process and trainable parameters of KV-Latent.}
    \label{fig:downsample}
\end{figure}

However, the introduction of RoPE failed this approach, as RoPE involves rotating pairs of channels at different frequencies. The specific implementation, which includes sparse matrix multiplication and the modern channel grouping approach found in GPT-NeoX~\citep{black-etal-2022-gpt}, is detailed in Appendix~\ref{sec:ropeimp}, in which uniform down-sampling is enough for weight initializing. Leveraging this methodology, an example of shrinking $d_{vo}$ by half and $d_{qk}$ by three quarters is described in Formula~\ref{for:downsample}.

\begin{equation}
\label{for:downsample}
\begin{split}
    \Tilde{W_Q^{(i)}} & = W_Q^{(i)}[:,::\!4]
    = W_{Q}[:, (i-1)d_{qk}:id_{qk}:\!4] \\
    \Tilde{W_K^{(i)}} & = W_K^{(i)}[:,::\!4]
    = W_{K}[:, (i-1)d_{qk}:id_{qk}:\!4] \\
    \Tilde{W_V^{(i)}} & = W_V^{(i)}[:,::\!2]
    = W_{V}[:, (i-1)d_{vo}:id_{vo}:\!2] \\
    \Tilde{W_O^{(i)}} & = W_O^{(i)}[::\!2,:]
    = W_{O}[(i-1)d_{vo}:id_{vo}:2, :]
\end{split}
\end{equation}

Recent works also apply the singular value decomposition (SVD) for down-sampling~\cite{DBLP:journals/corr/abs-2408-05646, zhang2024lorclowrankcompressionllms}, however, these methods faces major difficulties since the matrix multiplication does not satisfy the commutative property, which can't be applied here.

After the down-sampling step, we also hope to train FFNs to better let the model fit to it's modified attention, but not entirely forget what it has learnt in pre-training, so we apply Low Rank Adaption (LoRA, \citealp{DBLP:conf/iclr/HuSWALWWC22}) to FFNs' transformations, includes \texttt{Up}, \texttt{Down}, and \texttt{Gate} in a LLM which typically adapt Gated Linear Unit (GLU) as FFN. Figure~\ref{fig:downsample} describes our down-sampling and model building process.

\begin{figure*}
    \centering
    \includegraphics[width=1\linewidth]{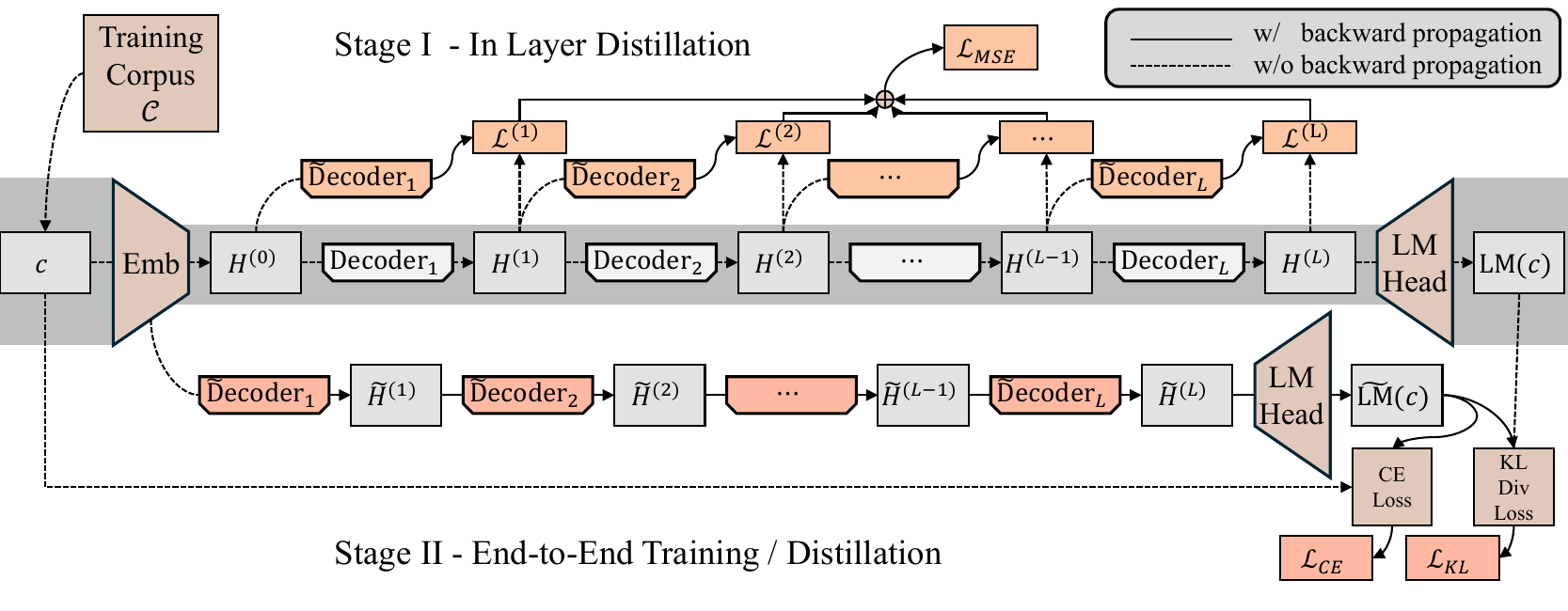}
    \caption{Dataflow of two stage training in KV-Latent.}
    \label{fig:2stage}
\end{figure*}

\subsubsection{Stage I  - In Layer Distillation}

In the first stage of training, we aim to maintain maximum consistency between the hidden states $H^{(l)}$ within two decoder layers. This approach ensures that we preserve the model’s initial capabilities to the greatest extent. To achieve this, we employ an in layer distillation method, depicted in Figure~\ref{fig:2stage}.

We define $H^{(l+1)}=\mathrm{Decoder}_l\left(H^{(l)}\right)$ as the operation of $l$-th Transformer decoder block of the initial model, and $\Tilde{\mathrm{D}}\mathrm{ecoder}_l(\cdot)$ as the modified version of it with a reduced $Q,K,V,O$ head size that utilize KV-Latent.
We first perform inference using the original $\mathrm{Decoder}(\cdot)$, retaining the intermediate hidden states $H_{\{0,1,\dots,L\}}$ between every two layers. 
For the $l$-th layer, we define three hidden states with identical shapes: $H^{(l)}_i, H^{(l)}_t, H^{(l)}_p$, as obtained from Formula~\ref{for:dist1},
\begin{equation}
\label{for:dist1}
\begin{split}
    H^{(l)}_i = H^{(l-1)},\, 
    & H^{(l)}_t = \mathrm{Decoder}_l(H^{(l)}_i) = H^{(l)} \\ 
    & H^{(l)}_p = \Tilde{\mathrm{D}}\mathrm{ecoder}_l(H^{(l)}_i)
\end{split}
\end{equation}
serves as the $i$nput, $t$arget, and $p$redicted hidden states.
We want to maximize the similarity between $H^{(l)}_t$ and $H^{(l)}_p$.
To achieve this, we use Mean Squared Error (MSE) loss. We define $W_{\mathrm{dec}}$ as the trainable weights of $\Tilde{\mathrm{D}}\mathrm{ecoder}(\cdot)$. Our optimization objective is described in Formula~\ref{for:dist1loss}.
\begin{equation}
\label{for:dist1loss}
    \min_{W_{\mathrm{dec}}} \frac1L \cdot \sum_{l=1}^{L}\frac{
        \mid \mid H^{(l)}_t - H^{(l)}_p \mid \mid_2
    }{x\cdot h}
\end{equation}

\subsubsection{Stage II - End-to-End Training / Distillation}

Despite performing intra-layer distillation, to apply KV-Latent on modern LLMs still faces challenges due to LLMs' depth. Even minor perturbations can be amplified layer by layer, potentially compromising their model's language capabilities. To address this, we need to train the model end-to-end. In this stage, we have two choices, Next-Token-Prediction (NTP) training and Distillation. We firstly define the original model $\mathrm{LM}(\cdot)$ and our KV-Latent model $\Tilde{\mathrm{LM}}(\cdot)$, and 
$\mathcal{C} = \{c_1, c_2, \dots, c_{\mid \mathcal{C} \mid}\}$ 
as the corpus we use for training, where 
$c_i = \{t_1,t_2,\dots,t_{\mid c_i \mid}\}$.

NTP training is part of the model’s pre-training and employs cross-entropy loss, described in Formula~\ref{for:CEloss}. It requires minimal resources, however, cross-entropy loss provides limited information.
\begin{equation}
\label{for:CEloss}
    \min_{W_{\mathrm{dec}}}
    \sum_{c\in \mathcal{C}} \sum_{x = 1}^{\mid c \mid - 1}
    \frac{
        \mathrm{CELoss}\left(
            \Tilde{\mathrm{LM}}(c)[x],\ c[x + 1]
        \right)
    }{\mid\!\mathcal{C}\!\mid \cdot \left(\mid\!c\!\mid - 1\right)}
\end{equation}

Distillation based on predicted probability distributions is commonly used for model recovery, this method relies on KL divergence loss, described in Formula~\ref{for:KLloss}. Distillation helps model to learn more with same amount of data. However, distillation involves an additional forward pass to compute the probability distributions and requires maintaining an extra set of parameters. 
\begin{equation}
\label{for:KLloss}
    \min_{W_{\mathrm{dec}}}
    \sum_{c\in \mathcal{C}} \sum_{x = 1}^{\mid c \mid}
    \frac{
        \mathrm{KLLoss}\left(
            \Tilde{\mathrm{LM}}(c)[x],\ \mathrm{LM}(c)[x]
        \right)
    }{\mid\!\mathcal{C}\!\mid \cdot \mid\!c\!\mid}
\end{equation}

\subsection{Frequency-aware RoPE for Variable Dimensions}

\def \One {1\!\!1}

\subsubsection{Motivation}
% We observed that RoPE failed on smaller $d_{qk}$, making the model training failed. To address this issue, we investigated the relative attention scaled score of RoPE-encoded vectors across different dimensions. As shown in Figure~\ref{fig:ropedef}, we finds out that, as the dimensionality decreases, the position encoding capability significantly diminishes. According to the \citeauthor{SU2024127063}’s analysis, when relative scores become negative, the encoding capability lost. RoPE remains acceptable on $64$ dimensions, however, at $32$ dimensions, a substantial number of negative scores emerge.

% \begin{figure}[t]
%     \centering
%     \includegraphics[width=1\linewidth]{Figures/FigureRoPEdefault.pdf}
%     \caption{The encoding ability of RoPE diminished with lower dimensional vectors, $\theta = 500000$.}
%     \label{fig:ropedef}
% \end{figure}

\begin{figure*}[ht]
\begin{minipage}[t]{0.24\linewidth}
    \centering
    \includegraphics[width=1\linewidth]{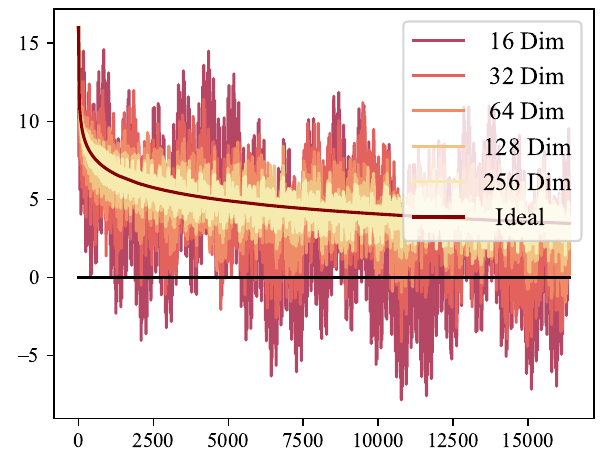}
    \caption{RoPE Diminish}
    \label{fig:ropedef}
\end{minipage}
\begin{minipage}[t]{0.24\linewidth}
    \centering
    \includegraphics[width=1\linewidth]{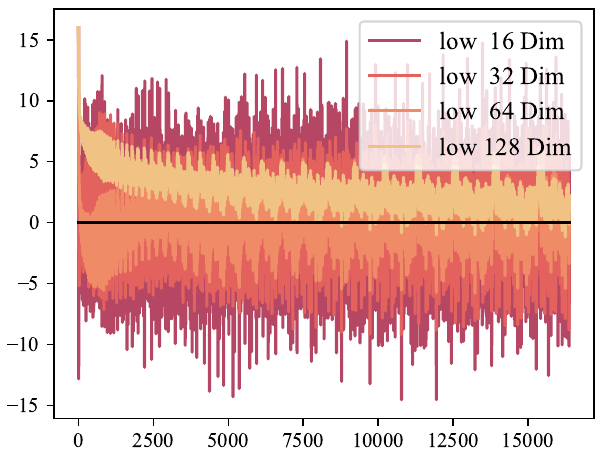}
    \caption{Lower dims only}
    \label{fig:ropelower}
\end{minipage}
\begin{minipage}[t]{0.24\linewidth}
    \centering
    \includegraphics[width=1\linewidth]{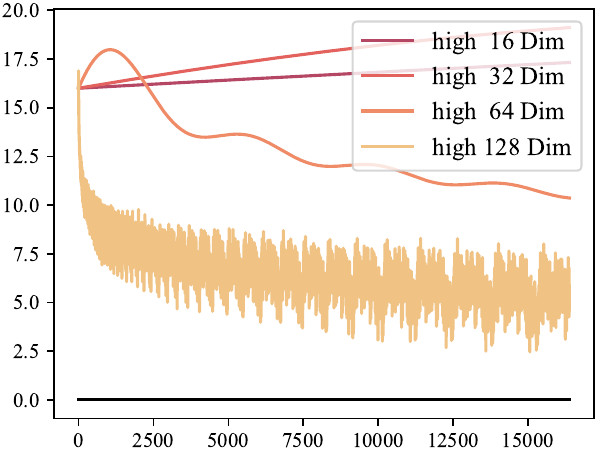}
    \caption{Higher dims only}
    \label{fig:ropehigher}
\end{minipage}
\begin{minipage}[t]{0.24\linewidth}
    \centering
    \includegraphics[width=1\linewidth]{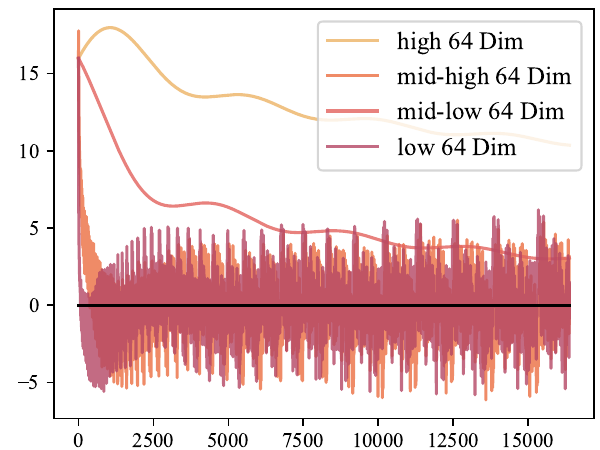}
    \caption{Quarter dims}
    \label{fig:ropequarter}
\end{minipage}
\end{figure*}

RoPE introduces positional information into the $Q$ and $K^\top$ components of the attention heads. 
In our preliminary experiments, we observed that RoPE exhibits significant numerical instability when applied to lower-dimensional vectors, as shown in Figure~\ref{fig:ropedef}. Specifically, when the dimension $d$ is smaller than $32$, the range of oscillation is comparable with intended attenuation, indicating the loss of positional encoding capability.
According to \citeauthor{SU2024127063}, vectors encoded by RoPE should maintain a certain degree of similarity with itself, even at large distances. 
% Specifically, the similarity between two encoded vectors of the same value should not be less than the similarity between two randomly generated vectors. In high-dimensional space, two random vectors are most likely to be orthogonal to each other, implying that the similarity between two identical vectors positioned at different locations should be greater than zero. 
We can measure this by using special vector $\One^d = (1,1,\cdots,1) \in \mathbb{R}^{d}$. We define $\mathrm{RoPE}_{\theta,d}(x)$ in Formula~\ref{for:ropeoones} as a representation of the similarity of two same vector across difference distance $x$, whose value ideally should always be positive to be more similar than two random vector.
\begin{equation}
\label{for:ropeoones}
    \mathrm{RoPE}_{\theta, d}(x)=\One_d \cdot \mathcal{R}_{\theta, \frac d2}(x) \cdot \One_d^\top
\end{equation}

\subsubsection{Pattern Finding}
We investigated the impact of different values of $d$ on $\mathrm{RoPE}_{\theta,d}$, as shown in Figure~\ref{fig:ropedef}. We observed smaller values of $d$ result in greater noise, along with an increased occurrence of negative values. We decomposed the vector by channels. Based on the $256$-dimensional case, Figures~\ref{fig:ropelower} and \ref{fig:ropehigher} illustrate the scenarios where low-numbered and high-numbered channels are retained, respectively, while Figure~\ref{fig:ropequarter} depicts the RoPE function for retaining different sets of $64$ consecutive channels (one-quarter of the total). Our findings suggest that the low-numbered channels of the RoPE function contribute the majority of the noise, while the high-numbered channels, despite a slower decay, remain relatively stable. Aligned with some previous works~\citep{blocntkaware, peng2024yarn}.

% We investigate the source of RoPE's instability. Initially, we retained different set of dimensions from a full $256$-dimensional vector and observed the trends in relative scores. When preserving only lower dimensions, RoPE quickly lost its encoding capability, as shown in Figure~\ref{fig:ropelower}. Conversely, when retaining higher dimensions, RoPE’s stability not only remained unchanged but even improved, as seen in Figure~\ref{fig:ropehigher}. However, the issue of ineffective attenuation arises. In fact, if we extend the observation range of relative positions, we find that long-range attenuation still exists, albeit at a slow pace. By separately preserving the first to forth 64-dimensional quarter-vector, in the Figure~\ref{fig:ropequarter}, we arrive at the same conclusion: low dimensions introduce noise but aid local attenuation, while high dimensions remain stable but suffer from slow attenuation.

\begin{figure*}[ht]
\begin{minipage}[t]{0.24\linewidth}
    \centering
    \includegraphics[width=1\linewidth]{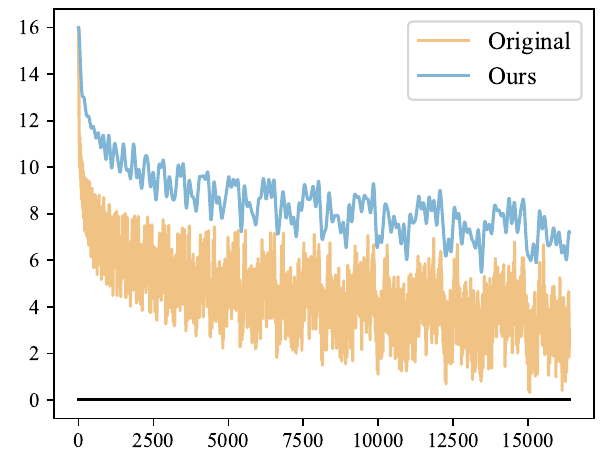}
    \caption{128 dim}
    \label{fig:ropemod128}
\end{minipage}
\begin{minipage}[t]{0.24\linewidth}
    \centering
    \includegraphics[width=1\linewidth]{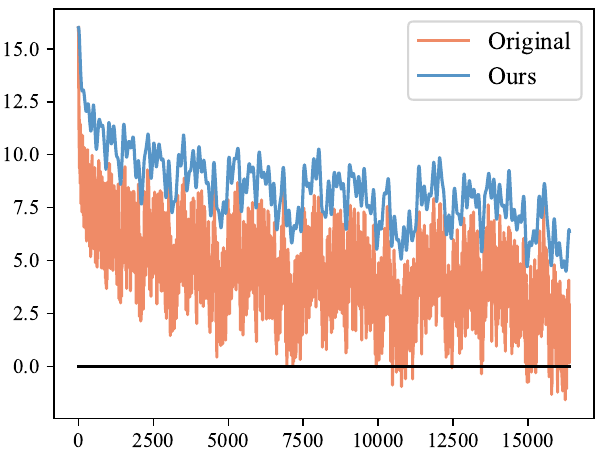}
    \caption{64 dim}
    \label{fig:ropemod64}
\end{minipage}
\begin{minipage}[t]{0.24\linewidth}
    \centering
    \includegraphics[width=1\linewidth]{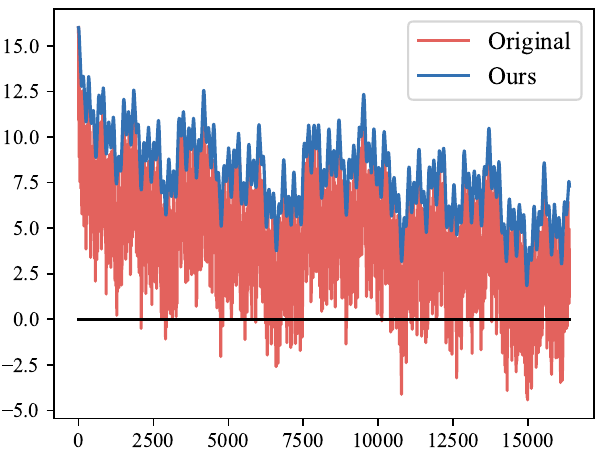}
    \caption{32 dim}
    \label{fig:ropemod32}
\end{minipage}
\begin{minipage}[t]{0.24\linewidth}
    \centering
    \includegraphics[width=1\linewidth]{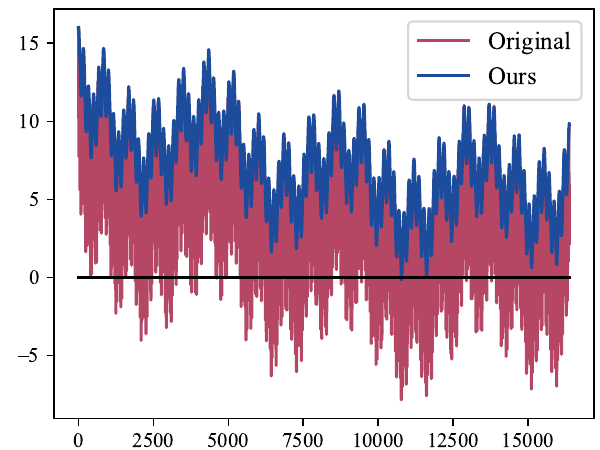}
    \caption{16 dim}
    \label{fig:ropemod16}
\end{minipage}
\end{figure*}

\subsubsection{Frequency-aware Modification}

% Building upon the conclusions from the previous section, we propose a modified sampling approach for RoPE. We found that lower-dimensional vectors tend to introduce noise. In Formula~\ref{for:ropedefault}, we notice that the low-dimensional components correspond to the high-frequency rotations in RoPE. Consequently, we need to reduce the dominance of these components and enhance the intensity of low-frequency ones. So we modify the original uniform frequency sampling by removing the highest-frequency portion and increasing the density of the low-frequency segment. This adjustment led us to the new frequency sampling, described in Formula~\ref{for:ropemod}. Modified results can be found in Figure~\ref{fig:ropemod128},~\ref{fig:ropemod64},~\ref{fig:ropemod32}, and~\ref{fig:ropemod16}, where we observe that although some noise remains, negative scores no longer occur. Moreover, we can find that our methods tend be become the upper limit of the original RoPE.

We implemented a frequency-aware modification strategy, which involves densifying the sampling of low-frequency rotations and avoiding high-frequency rotation sampling, as described in Formula~\ref{for:ropemod}, since that lower-numbered channels correspond to high-frequency rotations and higher-numbered channels correspond to low-frequency rotations The results, presented in Figures~\ref{fig:ropemod128},~\ref{fig:ropemod64},~\ref{fig:ropemod32}, and~\ref{fig:ropemod16}, demonstrate that our approach achieves enhanced stability while also reducing the occurrence of negative values.
\begin{equation}
\label{for:ropemod}
    \theta_j = \left\{ 
    \begin{aligned}
        & \theta^{-2\left(j-1 + d/8\right )/d}, \\
        & \quad \mathrm{for}\ j\in \left[1, 2, \dots, d/4 \right] \\
        & \theta^{-\left(j-1 + 3d/4\right )/d}, \\
        & \quad \mathrm{for}\ j\in \left[d/4 + 1, d/4 + 2, \dots, d/2 \right]
    \end{aligned}
    \right.
\end{equation}

\subsection{Effectiveness Analysis}

To explain why our method is effective, we present the following derivation. From the ideal curve in Figure~\ref{fig:ropedef}, it is evident that as $d$ increases, RoPE approaches a smooth decay curve. The calculation of this curve is given by Formula~\ref{for:ropeideal}, detailed in Appendix~\ref{sec:deriidealrope}. 

\begin{equation}
\label{for:ropeideal}
    \begin{aligned}
    \lim_{d\rightarrow +\infty} \frac1d \mathrm{RoPE}_{\theta, d}(x)
        = \int_{\log_\theta x -1}^{\log_\theta x} \cos(\theta^p)dp
    \end{aligned}
\end{equation}
\begin{figure}[htbp]
    \centering
    \includegraphics[width=1\linewidth]{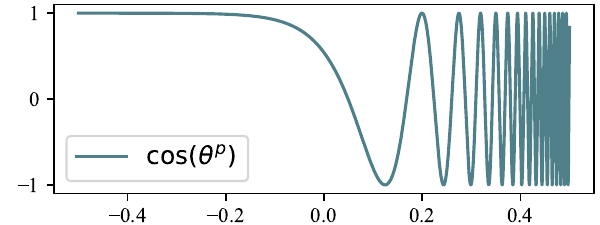}
    \caption{The $\cos(\theta^p)$ where $\theta=10000$}
    \label{fig:costheta}
\end{figure}

At this point, we transform the stability issue of RoPE into a problem of numerical approximation of an integral. Specifically, for the integral of the function $\cos (\theta ^p)$ over an interval of size $1$ (as shown in Figure~\ref{fig:costheta}), we approximate the solution by performing $d/2$ samples. The function exhibits sharp oscillations when $p$ takes larger values, and as $x$ increases, the integration window slides to the right, inevitably entering regions of these intense oscillations. Therefore, to accurately solve the integral, $d$ must be big enough for increased $x$. If $d$ is too small, the sampling interval may be shorter than the oscillation period, causing the numerical approximation to lose validity and introducing substantial noise.
We provided a code block to simulate this in Appendix~\ref{sec:ropecode}.

Furthermore, our modifications, by discarding certain sampling points on the right side, increased the overall sampling density while delaying the integration window's entry into the region of intense oscillations, enhancing the stability of the RoPE, thereby reducing noise amplitude. Additionally, the values of the extra sampling points are typically close to $1$, while the discarded points oscillate between $1$ and $-1$. As a result, the frequency-aware RoPE values are almost always greater than the original RoPE values, as detailed in Appendix~\ref{sec:proofmodlarge}.

\section{Experiments}
\subsection{Training}

We utilized FineWeb-edu~\citep{lozhkov2024fineweb-edu}, which is derived from FineWeb~\citep{DBLP:journals/corr/abs-2406-17557}, a web dataset based on open-access web pages consists 15 trillion token. We used a 1 billion token subset from FineWeb-edu, a common size also utilized by other well-known datasets, such as minipile~\citep{kaddour2023minipile,gao2020pile}. Our training hyperparameters are detailed in Appendix~\ref{sec:hyp}. We've conduct our training on a single node with 8 NVIDIA A100 80G SXM4 GPU.

Model wise, we've trained two versions of KV-Latent on LLaMA-3-8B(L3-8B), with $(d_{qk}, d_{vo}) = (64, 64)$ and $(16, 16)$ as a GQA examples, one version on LLaMA-2-7B(L2-7B) with $(d_{qk}, d_{vo}) = (64, 64)$, as an MHA example.

\begin{table*}[t]
\centering
\begin{tabular}{l|ccc||ccc|c||c|cc|cc}
    \toprule
    Model & $d_{qk}$ & $d_{vo}$ & Method & mmlu & obqa & arc & Avg
    & NIH & $s_{\mathrm{kv}}$ & $r_{\mathrm{s}}$ & $t_{\mathrm{ttft}}$ & $r_{\mathrm{t}}$ \\
    \midrule
    \multirow{4}{*}{L3-8B} 
        & 128 & 128 & Base    & 35.3 & 35.5 & 55.5 & 42.1 & 92\% & 491 & -    & 670 & -     \\
        & 64  & 64  & Train   & 35.0 & 35.1 & 53.8 & 41.3 & 92\% & 245 & ↓50\% & 622 & ↓8\%  \\
        & 64  & 64  & Distill & 31.0 & 29.1 & 39.1 & 33.1 & 94\% & 245 & ↓50\% & 622 & ↓8\%  \\
        & 16  & 16  & Train   & 31.0 & 29.5 & 38.5 & 33.0 & 6\%  & 64  & ↓87\% & 595 & ↓13\% \\
    \midrule
    \multirow{3}{*}{L2-7B} 
        & 128 & 128 & Base    & 28.9 & 29.4 & 30.7 & 29.7 & 32\% & 1966 & -    & 668 & -     \\
        & 64  & 64  & Train   & 28.1 & 29.3 & 27.5 & 28.3 & 24\% & 983 & ↓50\% & 573 & ↓17\% \\
        & 64  & 64  & Distill & 26.2 & 28.6 & 27.0 & 27.3 & 4\%  & 983 & ↓50\% & 573 & ↓17\% \\
    \bottomrule 
\end{tabular}
\caption{KV-Latent model's performance on benchmarks. NIH refers to \textit{Needle in haystack} testing result.}
\label{tab:metrics}
\end{table*}

% \begin{table*}[h]
% \centering
% \begin{tabular}{c|c|ccccccccccccccc}
%     \toprule
%     $d_{qk}$ & $128$ & $128$ & $128$ & $128$ & $64$ & $64$ & $64$ & $64$ & $32$ & $32$ & $32$ & $32$ & $16$ & $16$ & $16$ & $16$\\
%     $d_{vo}$ & $128$ & $64$ & $32$ & $16$ & $128$ & $64$ & $32$ & $16$ & $128$ & $64$ & $32$ & $16$ & $128$ & $64$ & $32$ & $16$\\
%     \midrule
%     Log PPL                 & -     & $2.80$ & $2.91$ & $3.01$ & $2.74$ & $2.74$ & $2.86$ & $3.12$ & $2.67$ & $2.79$ & $3.03$ & $3.29$ & $2.83$ & $3.00$ & $3.18$ & $3.78$ \\
%     \midrule
%     $t_{\mathrm{train}}$(H) & -     & $20.3$ & $19.6$ & $19.4$ & $20.1$ & $18.0$ & $17.5$ & $17.2$ & $19.1$ & $17.3$ & $16.6$ & $16.3$ & $19.1$ & $17.0$ & $16.3$ & $16.1$ \\
%     $t_{\mathrm{ttft}}$(ms) & -     & $425$ & $264$ & $260$ & $262$ & $256$ & $252$ & $246$ & $252$ & $245$ & $243$ & $239$ & $253$ & $246$ & $239$ & $238$ \\
%     $t_{\mathrm{mspt}}$(ms) & -     & $35.6$ & $37.2$ & $35.9$ & $35.9$ & $36.4$ & $35.7$ & $34.9$ & $35.1$ & $35.0$ & $35.2$ & $34.6$ & $35.2$ & $35.2$ & $35.0$ & $134.7$ \\
%     \midrule
%     $s_{\mathrm{kv}}$(KB)   & -     & $96$ & $80$ & $72$ & $96$ & $64$ & $48$ & $40$ & $80$ & $48$ & $32$ & $24$ & $72$ & $40$ & $24$ & $16$ \\
%     \bottomrule 
% \end{tabular}
% \caption{Impact of different $d_{qk}$ and $d_{vo}$.}
% \label{tab:rqkvo-metrics}
% \end{table*}

\begin{table*}[ht]
% \begin{minipage}[t]{0.745\linewidth}
    \centering
    \begin{tabular}{c|c||c|ccc|ccc|ccc}
        \toprule
        \multicolumn{2}{c||}{$d_{qk}$} & $128$  & $64$   & $32$   & $16$   
                             & $64$   & $32$   & $16$ 
                             & $128$  & $128$  & $128$  \\
        \multicolumn{2}{c||}{$d_{vo}$} & $128$  & $64$   & $32$   & $16$   
                             & $128$  & $128$  & $128$ 
                             & $64$   & $32$   & $16$   \\
        \midrule
        \multicolumn{2}{c||}{LogPPL}   & -      & $2.74$ & $3.03$ & $3.78$ 
                             & $2.47$ & $2.67$ & $2.83$ 
                             & $2.80$ & $2.91$ & $3.01$ \\
        \midrule
        $t_{\mathrm{train}}$ & hour & -      & $18.0$ & $16.6$ & $16.1$ 
                             & $20.1$ & $19.1$ & $19.1$ 
                             & $20.3$ & $19.6$ & $19.4$ \\
        $t_{\mathrm{ttft}}$ & ms & $303$  & $256$  & $243$  & $238$
                             & $262$  & $252$  & $260$  
                             & $296$  & $264$  & $238$  \\
        $t_{\mathrm{mspt}}$ & ms & $35.9$ & $36.4$ & $35.2$ & $34.7$ 
                             & $35.9$ & $35.1$ & $35.9$ 
                             & $34.9$ & $37.2$ & $34.7$ \\
        \midrule
        $s_{\mathrm{kv}}$ & MB & $256$  & $128$  & $64$   & $32$   
                             & $172$  & $160$  & $144$
                             & $172$  & $160$  & $144$  \\
        $n_{\mathrm{max}}$ & $10^6$ token & $0.40$ & $0.81$ & $1.63$ & $3.27$   
                             & $0.61$ & $0.65$ & $0.72$
                             & $0.61$ & $0.65$ & $0.72$ \\
        \bottomrule 
    \end{tabular}
    \caption{General performance of different $d_{qk}$ and $d_{vo}$.
    }
    \label{tab:rqkvo-metrics}
% \end{minipage}
% \begin{minipage}[t]{0.22\linewidth}
%     \centering
%     \begin{tabular}{cc|cc}
%         \toprule
%         $64$   & $32$   & $64$   & $16$   \\
%         $32$   & $64$   & $16$   & $64$   \\
%         \midrule
%         $2.86$ & $2.79$ & $3.12$ & $3.00$ \\
%         \midrule
%         $17.5$ & $17.3$ & $17.2$ & $17.0$ \\
%         $252$  & $245$  & $246$  & $246$  \\
%         $35.7$ & $35.0$ & $34.9$ & $35.2$ \\
%         \midrule
%         $96$   & $96$   & $80$   & $80$   \\
%         $1.09$ & $1.09$ & $1.31$ & $1.31$ \\
%         \bottomrule 
%     \end{tabular}
%     \caption{Same budget, high $d_{vo}$ gives better result.}
%     \label{tab:rvo-metrics}
% \end{minipage}
\end{table*}

% \begin{table*}[t]
% \begin{minipage}[t]{0.5\linewidth}
\begin{table}[t]
    \centering
    \begin{tabular}{l|ccccc}
        \toprule
        Rank & $16$ & $32$ & $64$ & $128$ & $256$\\
        \midrule
        $t_{\mathrm{train}}$(H) & $16.9$ & $16.8$ & $17.1$ & $17.5$ & $18.0$ \\
        LogPPL               & $2.49$ & $2.47$ & $2.46$ & $2.46$ & $2.45$ \\
        \bottomrule 
    \end{tabular}
    \caption{KV-Latent with different LoRA rank.}
    \label{tab:lora-metrics}
\end{table}
\begin{table}[t]
% \end{minipage}
% \begin{minipage}[t]{0.48\linewidth}
    \centering
    \begin{tabular}{l|ccc}
        \toprule
        Method                    & Log PPL & \multicolumn{2}{c}{Avg $s_{\mathrm{kv}}$} \\
        \midrule
        KV-L                 & $2.509$ & $128$ & ↓50\% \\
        KV-L + PyI & $2.499$ & $64$  & ↓75\% \\
        \bottomrule 
    \end{tabular}
    \caption{KV-Latent(KV-L) with PyrimaidInfer(PyI).}
    \label{tab:kvlpyi}
% \end{minipage}
% \end{table*}
\end{table}

\subsection{Evaluation}

We conducted tests on the KV-Latent model from two perspectives: performance and efficiency. For performance, we used 0-shot MMLU~\cite{DBLP:conf/iclr/HendrycksBBZMSS21}, OBQA~\cite{mihaylov-etal-2018-suit}, and AI2ARC~\cite{DBLP:journals/corr/abs-1803-05457}, as benches. Additionally, we performed a \textit{needle in a haystack} (NIH) test to assess the ability of information retrieval. We put a short sentence randomly in a 3,840 tokens context, and check whether the model could retell it, repeat 50 times and calculate the pass ratio. Regarding efficiency, we measured the KV Cache size $s_{\mathrm{kv}}$ (MB) during the NIH experiment and the latency to the first token $t_{\mathrm{ttft}}$ (ms). We've also calculate the improve ratio $r_{\mathrm{s}}$ and $r_{\mathrm{t}}$. Results are shown in Table~\ref{tab:metrics} with several key observations. Firstly, KV-Latent allows the model to achieve satisfactory performance while reducing the KV Cache size. Secondly, despite distillation transfer more information, the limited training volume unables it to fully recover model's proficiency. Thirdly, when $d_{qk}=d_{vo}=16$, the model’s performance failed to be recovered, suggesting a lower bound of KV Cache size. Lastly, LLaMA2, which does not utilize GQA, relatively outperforms LLaMA3 when trained on fewer tokens, indicating that for models already trained with GQA, adopting KV-Latent presents additional challenges.

\subsection{Impact of Parameter Selection}
\label{sec:iops}
We investigated the impact of different $d_{qk}$, $d_{vo}$, and LoRA rank, on model's performance. We conducted experiments using the LLaMA-3-8B base model and trained multiple versions of KV-Latent with varying configurations. By default, we set $d_{qk} = d_{vo} = 64$ and LoRA rank$= 256$. For efficiency-related tests, we generated 256 tokens based on a context length of 2048, repeating the process 15 times and averaging the results.

\subsubsection{Combinations of QK and VO Head Size}

We test different combinations $d_{qk}$ and $d_{vo}$ on model performance and efficiency. We encompass three aspects: logarithmic perplexity (log PPL), reflecting model’s language modeling ability; training speed $t_{\mathrm{train}}$, measuring the training efficiency; and inference speed, includes time to the first token $t_{\mathrm{ttft}}$, and millisecond per new token $t_{\mathrm{mspt}}$. In terms of space KV Cache size $s_{\mathrm{kv}}$ for the 4,000 token length sequence under \texttt{bfloat16}. For a more intuitive representation, we calculated the maximum KV Cache size $n_{\mathrm{max}}$ supported with 60GB of memory, as an 80GB compute card scenario, excludes approximately 15GB for model parameters and 5GB as buffer. Results are shown in Table~\ref{tab:rqkvo-metrics}. 

We find that, firstly, the efficiency related to the KV Cache aligns with it's size: the smaller the overall volume, the faster both pre-filling and generation speeds. However, when comparing the results of reducing $d_{qk}$ versus $d_{vo}$, in Table~\ref{tab:rvo-metrics} Appendix~\ref{sec:rvo-metrics}, we noticed that allocating more resources to $d_{vo}$ consistently yields better efficiency and effectiveness, suggesting that \texttt{Key}s carry less essential information than \texttt{Value}s within the KV Cache, making them more amenable to compression.

\subsubsection{LoRA Rank}

LoRA rank may impact KV-Latent's performance and efficiency. We focused on evaluating training efficiency and log PPL since LoRA possess no extra cost in inference. Shown in Table~\ref{tab:lora-metrics}, increasing the rank corresponds to increase in training time. However, we noticed that the change in log PPL is less significant. 
It’s important to note that LoRA rank might have a more substantial effect in larger-scale training scenarios.

\subsubsection{Cross-method Feasibility}

In terms of compatibility with other methods, KV-Latent works well with Head-Level, as evidenced by the tests on LLaMA-3. It is also compatible with Layer-Level, although the higher training resource requirements. Finally, our method is also compatible with Token-Level. Table~\ref{tab:kvlpyi} shows the results when used in conjunction with PyramidInfer with 50\% compress rate, as one of the popular token-level reduction methods, proving our statement.
KV-Latent is orthogonal with all main-stream methods.

% \section{Limitations and Future Work}
% \subsection{Training Requirement}
% The paradigm of KV-Latent, even though it requires significantly less training data than pretraining, still incurs substantial costs. It remains distant from the ideal low-rank and low-resource training scenario. Exploring better dimension selection or pruning algorithms to reduce the required additional training cost could be a promising avenue for future research.

% \subsection{Multi-Dimension Integration}
% Our work demonstrates the feasibility of reducing the dimensionality of attention vectors. However, if we can take it a step further and allow different tokens to use varying attention vector sizes while achieving alignment across different dimensions, we could create a more flexible LLM and KV Cache. Exploring the possibility of cross-dimensional alignment and KV Cache with anisotropy token could be one of the future research directions.

\section{Conclusion}
We propose KV-Latent, a paradigm that directly reduces the model’s attention head dimensionally, thus KV Cache size, through a two-step training process. This approach achieves cache reduction and enhancing inference speed while using only a small number of additional tokens for training. We have demonstrated that decoupling the relationships of $n_h \cdot d_h = d$ and $d_h=d_{qk}=d_{vo}$ is feasible. Notably, we found that $d_{vo}$ has a greater impact on model performance, revealing an information imbalance between values and keys within the KV Cache. Finally, by modifying the frequency sampling method, we enhance RoPE’s stability while preserving its attenuation properties. Out work may contribute to the study of optimizing model structure.

\section*{Limitations}
Currently, we are unable to perform a direct comparison with certain related methods, such as Cross Layer Attention (CLA) mentioned earlier. Our approach only requires a limited amount of additional training, the outcome is still based on an existing model. Comparing it to CLA, which necessitates complete retraining, would be unfair and would exaggerate the effectiveness of our method, rendering the comparison meaningless.

Another potential direction for extension is the integration of SVD into the KV-Latent, which could provide the model with additional initial information. However, due to the inherent properties of RoPE and matrix multiplication, while this remains a possibility, it is overall highly challenging and would require substantial modifications to the model.

Additionally, our paper's discussion predominantly focuses on the pre-training phase of the model, without delving deeply into the aspects of Supervised Fine-Tuning and Reinforcement Learning from Human Feedback and their potential impacts. But currently, there is no evidence to suggest that our method presents any compatibility issues with SFT or RLHF.

Finally, our method aims to accelerate the inference of LLM without introducing security concerns greater than those inherent to the LLM itself.

\bibliography{custom}

\appendix

\section{Training Hyper-parameters}
\label{sec:hyp}
Due to computing resource limitations, we can only use a limited amount of tokens for some training.

\begin{table}[htbp]
\centering
\begin{tabular}{l|rr}
    \toprule
    Hyperparameter & \multicolumn{2}{c}{Value} \\
    \midrule
    $(d_{qk}, d_{vo})$ & \multicolumn{2}{c}{$(64, 64)$, $(16, 16)$} \\
    LoRA Rank & \multicolumn{2}{c}{$256$} \\
    LoRA $\alpha$ & \multicolumn{2}{c}{$512$} \\
    \midrule
    Batch Size & \multicolumn{2}{c}{$8$} \\
    Max Seq. Length & \multicolumn{2}{c}{$4096$} \\
    \multirow{2}{*}{Learning Rate}
        & 2e-5 & (Training) \\
        & 2e-7 & (Distillation) \\
    \multirow{4}{*}{Token Used} 
        & $0.1$B  & (Stage I) \\
        & $0.25$B & (Stage II Distill) \\
        & $1$B    & (Stage II Train) \\
        & $0.25$B & (Param Selection) \\
    \midrule
    Optimizer & \multicolumn{2}{c}{AdamW} \\
    Adam $\epsilon$ & \multicolumn{2}{c}{2e-4} \\
    Adam $\beta$s & \multicolumn{2}{c}{$(0.9,\,0.999)$} \\
    Weight Decay  & \multicolumn{2}{c}{$0.01$} \\
    \midrule
    Scheduler & \multicolumn{2}{c}{Cosine Annealing} \\
    \bottomrule 
\end{tabular}
\caption{Hyperparameters used for training.}
\label{tab:hyperparameters}
\end{table}

\section{Other Combinations of QK\&VO Heads}
\label{sec:rvo-metrics}
\begin{table}[htbp]
    \centering
    \begin{tabular}{l|cc|cc}
        \toprule
        $d_{qk}$             & $64$   & $32$   & $64$   & $16$   \\
        $d_{vo}$             & $32$   & $64$   & $16$   & $64$   \\
        \midrule
        LogPPL               & $2.86$ & $2.79$ & $3.12$ & $3.00$ \\
        \midrule
        $t_{\mathrm{train}}$ & $17.5$ & $17.3$ & $17.2$ & $17.0$ \\
        $t_{\mathrm{ttft}}$  & $252$  & $245$  & $246$  & $246$  \\
        $t_{\mathrm{mspt}}$  & $35.7$ & $35.0$ & $34.9$ & $35.2$ \\
        \midrule
        $s_{\mathrm{kv}}$    & $96$   & $96$   & $80$   & $80$   \\
        $n_{\mathrm{max}}$   &$1.09$ & $1.09$ & $1.31$ & $1.31$ \\
        \bottomrule 
    \end{tabular}
    \caption{Same budget, high $d_{vo}$ gives better result.}
    \label{tab:rvo-metrics}
\end{table}

\newpage
\section{RoPE Implementations}
\label{sec:ropeimp}

According to Formula~\ref{for:ropedefault}, RoPE is represented by a sparse matrix, and its computation in the sparse state is described by Formula~\ref{for:ropesparse}. 
\begin{equation}
\label{for:ropesparse}
\begin{split}
    & \mathcal{R}_{\theta, \frac{d}{2}}(x)y = \\
    & \begin{pmatrix}
             y_1\\ y_2 \\ y_3 \\ y_4 \\ \vdots \\ y_{d-1} \\ y_d
    \end{pmatrix} \! \otimes \! 
    \begin{bmatrix}
             \cos x\theta_1\\ \cos x\theta_1 \\ \cos x\theta_2\\ \cos x\theta_2 \\ \vdots \\ \cos x\theta_{\delta} \\ \cos x\theta_{\delta}
    \end{bmatrix} \! +\! 
    \begin{pmatrix}
             -y_2\\ y_1 \\ -y_4 \\ y_3 \\ \vdots \\ -y_d \\ y_{d-1}
    \end{pmatrix} \! \otimes\! 
    \begin{bmatrix}
             \sin x\theta_1\\ \sin x\theta_1 \\ \sin x\theta_2\\ \sin x\theta_2 \\ \vdots \\ \sin x\theta_{\delta} \\ \sin x\theta_{\delta}
    \end{bmatrix}
\end{split}
\end{equation}
In default RoPE strategy, each dimension of a head is paired, or shares the same $\theta_j$, with its neighbor, $2j$-th dimension is paired with $2j+1$-th mathematically. However, in popular frameworks like Transformers~\citep{wolf-etal-2020-transformers}, this process is achieved using Formula~\ref{for:ropesparsemixed}, which is firstly proposed in GPT-NeoX~\citep{black-etal-2022-gpt}. 
\begin{equation}
\label{for:ropesparsemixed}
\begin{split}
    & \mathcal{R}_{\theta, \frac{d}{2}}(x)y = \\
    & \begin{pmatrix}
             y_1 \\ y_2 \\ \vdots \\ y_{\delta} \\ 
             y_{\delta + 1} \\ y_{\delta + 2} \\ \vdots \\ y_d
    \end{pmatrix} \! \otimes \!
    \begin{bmatrix}
             \cos x\theta_1 \\ \cos x\theta_2 \\ \vdots \\ \cos x\theta_{\delta} \\ 
             \cos x\theta_1 \\ \cos x\theta_2 \\ \vdots \\ \cos x\theta_{\delta}
    \end{bmatrix} \! + \!
    \begin{pmatrix}
             -y_{\delta + 1} \\ -y_{\delta + 2} \\ \vdots \\ -y_d \\
             y_1 \\ y_2 \\ \vdots \\ y_{\delta}
    \end{pmatrix} \! \otimes \!
    \begin{bmatrix}
             \sin x\theta_1 \\ \sin x\theta_2 \\ \vdots \\ \sin x\theta_{\delta} \\ 
             \sin x\theta_1 \\ \sin x\theta_2 \\ \vdots \\ \sin x\theta_{\delta}
    \end{bmatrix}
\end{split}
\end{equation}
The actual RoPE matrix involved in computations pairs the dimensions $j$ and $j+\frac d2$. Consequently, we need to simultaneously select dimensions $j$ and $j+\frac d2$. To address this, we employ uniform sampling, which effectively satisfies this characteristic.

\onecolumn

\section{Detailed Formulas}

\subsection{Derivation of Ideal RoPE Curve}
\label{sec:deriidealrope}
\begin{equation*}
    \begin{aligned}
    \lim_{d\rightarrow +\infty} \frac1d \mathrm{RoPE}_d(x)
        &=\lim_{d\rightarrow +\infty} \frac1d \left(\One_d \cdot \mathcal{R}_{\theta, \frac{d}{2}}(x)\cdot {\One_d}^\top\right) \\
        &= \lim_{d\rightarrow +\infty} \frac1d \sum_{j=1}^{d/2} \One_2 \cdot \begin{pmatrix} 
            \cos(x\theta^{-2j/d}) & \sin(x\theta^{-2j/d}) \\
            -\sin(x\theta^{-2j/d}) & \cos(x\theta^{-2j/d})
        \end{pmatrix} \cdot {\One_2}^\top \\
        &= \lim_{d\rightarrow +\infty} \sum_{j=1}^{d/2} \cos(x\theta^{-2j/d}) \frac2d \\
        &=\lim_{d/2\rightarrow +\infty} \sum_{j=1}^{d/2} \cos(x\theta^{-2j/d}) \frac2d \\
        & = \int_0^1 \cos(x\theta^{-p}) dp \\
        & = \int_0^1 \cos(\theta^{\log_\theta x-p}) dp \\
        & = \int_{\log_\theta x -1}^{\log_\theta x} \cos(\theta^p)dp
    \end{aligned}
\end{equation*}

\subsection{Proof of Frequency-aware RoPE is Always Larger in Value}
\label{sec:proofmodlarge}
Firstly,
\begin{equation*}
    \begin{aligned}
    & \left\{
        \begin{aligned}
        \mathrm{RoPE} &=  \sum_{j=1}^{d/2} \cos(x\theta^{-2j/d}) \frac2d & (1) \\
        \mathrm{RoPE_{Mod}} &= \sum_{j=d/8 + 1}^{3d/8} \cos(x\theta^{-2j/d}) \frac2d  + \sum_{j=3d/8 + 1}^{d/2} \cos(x\theta^{-2j/d}) \frac2d & (2) \\
        \end{aligned}
    \right.  \\
    \Longrightarrow &
    \mathrm{RoPE_{Mod}} - \mathrm{RoPE} =
    \sum_{j=3d/8 + 1}^{d/2} \cos(x\theta^{-2j/d}) \frac2d - \sum_{j=1}^{d/8} \cos(x\theta^{-2j/d}) \frac2d
    \end{aligned}
\end{equation*}
And
\begin{equation*}
    \begin{aligned}
    j \in (\frac{3d}{8} + 1, \frac{d}{2})
    & \Rightarrow -\frac{2j}{d} \in (-1, -\frac34,) \\
    & \Rightarrow x\theta^{-2j/d} \approx 0
    \quad (\theta \gg x)
    \\
    & \Rightarrow \cos(x\theta^{-2j/d}) \approx 1
    \end{aligned}
\end{equation*}
Moreover
\begin{equation*}
    \cos(x\theta^{-2j/d}) \leq 1
\end{equation*}
So
\begin{equation*}
   \mathrm{RoPE_{Mod}} - \mathrm{RoPE} > 0 
\end{equation*}

\newpage
\section{RoPE Decay Curve Drawer Code}
\label{sec:ropecode}
A code piece to generate the rope decay curve with python, pytorch, and matplotlib. You can tune \texttt{theta} and \texttt{d} to see how $\mathrm{RoPE}_{\theta, d}(x)$ is affected by it's two hyper-parameters. Commonly, set $d=64$ or $128$ to get the curve of most common models like LLaMAs~\citep{dubey2024llama3herdmodels}. Or set $d$ to a very large value, i.e. $100000$, to draw the ideal curve.

\begin{lstlisting}[style=mypython]
import torch
from tqdm import tqdm
import matplotlib.pyplot as plt

device = torch.device("cuda" if torch.cuda.is_available() else "cpu")

theta = 10000.   # RoPE theta.
d = 100000       # Head dim.
steps = torch.arange(0, 1, 1 / d, device=device)

vals = []
MAX_POS_ID = 8192
for pos in tqdm(range(MAX_POS_ID)):
    with torch.no_grad():
        val = (((theta ** -steps) * pos).cos() / d).sum(dim=-1)
    vals.append(val.cpu().item())

plt.plot(torch.arange(MAX_POS_ID), vals)
plt.show()
\end{lstlisting}

\end{document}